\documentclass[letterpaper, 10 pt, conference]{ieeeconf}  
\pdfoutput=1

\usepackage{amsmath,amsfonts}
\usepackage{algorithmic}
\usepackage{algorithm}
\usepackage{array}
\usepackage[caption=false,font=normalsize,labelfont=sf,textfont=sf]{subfig}
\usepackage{textcomp}
\usepackage{stfloats}
\usepackage{url}
\usepackage{verbatim}
\usepackage{graphicx}
\usepackage{cite}
\usepackage{ulem} 
\usepackage{color}
\usepackage{setspace}
\usepackage{multirow}
\usepackage{tabularx}
\usepackage{booktabs}
\usepackage{ulem}
\IEEEoverridecommandlockouts                              

\title{\LARGE \bf
InterTracker: Discovering and  Tracking General Objects Interacting with Hands in the Wild
}

\author{Yanyan Shao$^{1}$, Qi Ye$^{2}$, Wenhan Luo$^{3}$, Kaihao Zhang$^{4}$, Jiming Chen$^{1, 2}$  
\thanks{
*This work is supported by NSFC 62088101 Autonomous Intelligent Unmanned Systems and NSFC 62103372, the Fundamental Research Funds for the Central Universities (226-2022-00107, 226-2023-00111). (Corresponding author: Qi Ye)
}
\thanks{$^{1}$College of Computer Science and Technology, Zhejiang University of Technology, Hangzhou, 310023, China.}
\thanks{$^{2}$College of Control Science and Engineering, Zhejiang University, Hangzhou, 310027, China.}
\thanks{$^{3}$School of Cyber Science and Technology, Sun Yat-sen University, Shenzhen, 518107, China.}
\thanks{$^{4}$College of Engineering, Computing \& Cybernetics, Australian National University, Canberra, ACT 2601, Australia. }}

\begin{document}

\newcommand{\reflabel}{dummy} 


\newcommand{\seclabel}[1]{\label{sec:\reflabel-#1}}
\newcommand{\secref}[2][\reflabel]{Section~\ref{sec:#1-#2}}
\newcommand{\Secref}[2][\reflabel]{Section~\ref{sec:#1-#2}}
\newcommand{\secrefs}[3][\reflabel]{Sections~\ref{sec:#1-#2} and~\ref{sec:#1-#3}}

\newcommand{\eqlabel}[1]{\label{eq:\reflabel-#1}}
\renewcommand{\eqref}[2][\reflabel]{(\ref{eq:#1-#2})}
\newcommand{\Eqref}[2][\reflabel]{(\ref{eq:#1-#2})}
\newcommand{\eqrefs}[3][\reflabel]{(\ref{eq:#1-#2}) and~(\ref{eq:#1-#3})}

\newcommand{\figlabel}[2][\reflabel]{\label{fig:#1-#2}}
\newcommand{\figref}[2][\reflabel]{Fig.~\ref{fig:#1-#2}}
\newcommand{\Figref}[2][\reflabel]{Fig.~\ref{fig:#1-#2}}
\newcommand{\figsref}[3][\reflabel]{Figs.~\ref{fig:#1-#2} and~\ref{fig:#1-#3}}
\newcommand{\Figsref}[3][\reflabel]{Figs.~\ref{fig:#1-#2} and~\ref{fig:#1-#3}}

\newcommand{\tablelabel}[2][\reflabel]{\label{table:#1-#2}}
\newcommand{\tableref}[2][\reflabel]{Table~\ref{table:#1-#2}}
\newcommand{\Tableref}[2][\reflabel]{Table~\ref{table:#1-#2}}
\newcommand{\etal}{et al.}
\newcommand{\eg}{e.g.}
\newcommand{\ie}{i.e. }
\newcommand{\etc}{etc. }

\def\bfmu{\mbox{\boldmath$\mu$}}
\def\bftau{\mbox{\boldmath$\tau$}}
\def\bftheta{\mbox{\boldmath$\theta$}}
\def\bfdelta{\mbox{\boldmath$\delta$}}
\def\bfphi{\mbox{\boldmath$\phi$}}
\def\bfpsi{\mbox{\boldmath$\psi$}}
\def\bfeta{\mbox{\boldmath$\eta$}}
\def\bfnabla{\mbox{\boldmath$\nabla$}}
\def\bfGamma{\mbox{\boldmath$\Gamma$}}

%
%


\newcommand{\R}{\mathbb{R}}

\newcommand{\be}{\begin{equation}}
\newcommand{\ee}{\end{equation}}
\newcommand{\yq}[1]{\textcolor{red}{\textbf{yq: }\xspace#1}\xspace}

\maketitle
\thispagestyle{empty}
\pagestyle{empty}


\begin{abstract}
Understanding human interaction with objects is an important research topic for embodied Artificial Intelligence and identifying the objects that humans are interacting with is a primary problem for interaction understanding. Existing methods rely on frame-based detectors to locate interacting objects. However, this approach is subjected to heavy occlusions, background clutter, and distracting objects.
To address the limitations, in this paper, we propose to leverage spatio-temporal information of hand-object interaction to track interactive objects under these challenging cases. Without prior knowledge of the general objects to be tracked like object tracking problems,  we first utilize the spatial relation between hands and objects to adaptively discover the interacting objects from the scene. Second, the consistency and continuity of the appearance of objects between successive frames are exploited to track the objects. With this tracking formulation, our method also benefits from training on large-scale general object-tracking datasets.  
We further curate a video-level hand-object interaction dataset for testing and evaluation from 100DOH.
The quantitative results demonstrate that our proposed method outperforms the state-of-the-art methods. Specifically, in scenes with continuous interaction with different objects, we achieve an impressive improvement of about $10\%$ as evaluated using the Average Precision (AP) metric.
Our qualitative findings also illustrate that our method can produce more continuous trajectories for interacting objects.
 
\end{abstract}

\section{INTRODUCTION}
Understanding human interaction with objects is an important research topic for embodied Artificial Intelligence. It enables robots to interact and collaborate with humans. The distilled knowledge of the interaction can also aid robots to learn autonomous skills like grasping \cite{karunratanakul2020grasping, jiang2021hand} and manipulation \cite{2022-eccv-dexmv, 2021-icra-learning}.

\begin{figure}
\centering 
 	\includegraphics[scale=1.4]{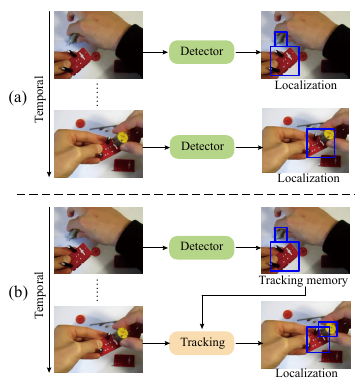}
	\caption{
 Comparison of generating interacting object trajectories by frame-to-frame detection (a) and our proposed TIO (b). We propose to track interacting objects in the sequence by utilizing historical detection results as tracking memory. Our proposed TIO can better cope with heavy occlusion and background clutter, thus enabling continuous localization. 
 }
 	\vspace{-20pt}
	\label{fig-tro}
\end{figure}

To understand the interaction, identifying the objects that humans are interacting with is a primary problem. There has been a great deal of excellent work \cite{2019-cvpr-contextual,2015-iccv-lending, 2018-tip-Joint, 2018-cvpr-lifestyle} focusing on hand perception and localization during the hand object interaction, while localizing general objects interacting with hands without the prior like templates over time in unconstrained scenes remains an under-explored area. Fouhey \etal \cite{100doh} use a hand-object detector to localize objects in interaction in each frame.  
However, during the interaction,  objects are frequently occluded by the hands and the presence of background clutters and distracting objects interferes with the estimation of object interaction states (contact / no contact). Both lead to missing alarms or inaccurate bounding box predictions.
Figure~\ref{fig-tro} shows an example of two hands interacting with two objects, respectively. The frame-based detector can detect and locate the interacting objects well but fails to distinguish between them and detect the small one when overlap happens.

To address the limitations of the frame-based detectors for general objects, in this paper, we propose to leverage spatio-temporal information of hand-object interaction to track interactive objects under severe occlusion and other interfering factors. The spatial relation between hands and objects is first utilized to adaptively discover the interacting objects from the scene to be tracked. Second, the consistency and continuity of the appearance of objects between successive frames are exploited to track the objects.

Different from the general object tracking task \cite{zhou2022apptracker, luo2019end, luo2021multiple}, the objects to be tracked are not available in advance in our problem. We hypothesize that the gesture of the hand implicitly indicates the geometry of the interacting object and propose an interaction detection module based on the hypothesis to identify the correct objects to be tracked in cluttered scenes. 
This module comprises two branches: a detection branch to detect the hands and candidate objects, and an interaction branch to estimate the object's location interacting with each detected hand based on its hand features. The object to be tracked is chosen according to the compatibility of locations of detected object candidates with the location from the interaction branch.

Modeling the interacting motion can better cope with the occlusion and deformation of interactive objects. 
However, building such a model faces the problem of data scarcity, due to the lack of video-level hand-object datasets that annotate bounding boxes of interacting objects. 
Inspired by the success of the Siamese-based trackers \cite{siamrpn++, SiamCAR}, we treat the motion modeling as a similarity matching problem between the template and the search region. For each interacting object, the historical state can act as the template to perform information embedding with the current frame. 
Considering that the motion is smooth, we determine the search region for similarity matching within the current frame based on historical position.
Instead of searching over the whole image, this can significantly improve computational efficiency.
With such a design, the motion modeling module can be trained to utilize a wide range of tracking datasets, which obviates the need for laborious and time-consuming video-level annotation of interactive objects.

The motion modeling module along with the detection module forms a novel framework to adaptively discover interacting objects and keep track of them in videos. 
To measure our method (TIO), we collect 2511 sequences with partial annotations from 100DOH \cite{100doh} for testing and evaluation, namely DOH-Trk. 
Qualitative and quantitative results show that our method surpasses the state-of-the-art method 100DOH \cite{100doh} and produces continuous trajectories for interacting objects.

The main contributions are as follows: 

\begin{itemize}
\itemsep 0pt
    \item We propose a novel object tracking framework leveraging spatio-temporal information during hand-object interaction to track general objects without priors in the wild.
    
    
    \item We curate a video-level dataset for the evaluation of tracking general objects during interaction based on 100DOH.

    \item Compared with the state-of-the-art method, our proposed approach significantly improves the localization of interacting objects.      
\end{itemize}

\section{Related work}
Our work focuses on improving the localization of objects interacting with hands through motion modeling in image sequences, which is fundamental to understanding human-object interactions at the Internet scale. In the following, we mainly review work relevant to hand and object detection.

Understanding hand-object interaction is an important and challenging task in computer vision and robotics \cite{9839629, 9810792}. 
There are numerous works on studying hand-object interaction from visual images, such as hand-object pose estimation \cite{liu2021semi, cai2019exploiting},
joint hand-object 3d reconstruction \cite{Chen2020JointH3, 2021-cvpr-reconstructing},
grasps generation \cite{jiang2021hand, karunratanakul2020grasping}
and learning to manipulate objects from visual demonstrations \cite{2022-eccv-dexmv, 2022-cvpr-human}.
The detection of hands and interactive objects in images as a basic task can further advance the study of these higher-level tasks.
Many researchers have devoted to hand detection \cite{2015-iccv-lending,2019-cvpr-contextual, 2018-cvpr-lifestyle, li2013pixel, 2022-cvpr-forward}.
Bambach \textit{et al.} \cite{2015-iccv-lending} propose a first-view collected hand dataset, and builds a CNN network for detecting and segmenting hands. To extend the detection method to unrestricted scenes, \cite{2019-cvpr-contextual, 2018-cvpr-lifestyle} enrich the hand detection dataset by sampling frames from YouTube videos. 
\cite{2020-nips-detecting} investigates to localize hands and recognize their psychical contact state,  \cite{2022-cvpr-forward} jointly detects and tracks hands online in unconstrained videos.
Since good progress has been made in hand detection techniques, in this work we aim to improve the localization of interacting objects.

For interacting objects, researchers have studied from different perspectives. 
\cite{misra2017red, naeem2021learning, nagarajan2018attributes}
learn state properties of objects at the image level, 
\cite{fathi2013modeling, liu2017jointly}
build explicit models to explore object state changes and associated state modifying actions in videos and \cite{2020-eccv-forecasting, 2021-cvpr-anticipative} analyze scene information for the prediction of future interactive objects.
However, the detection of interacting objects has been hampered by the lack of datasets that provide both hand and object annotations.
Shan \textit{et al.} \cite{100doh} propose a dataset that is collected from Internet videos and annotated the bounding boxes of both hands and objects, as well as the sides and contact states of the hands. They build a detector specifically for hand-object detection and demonstrate that the hand data obtained by the detector can be beneficial for hand mesh reconstruction. \cite{2022-eccv-fine} proposes an egocentric hand-object segmentation dataset for more accurate hand-object localization. They demonstrate that perceiving the position of the hand and object can greatly improve the performance of hand states classification and activity recognition.
\cite{2021-nips-cohesiv} introduces a weakly supervised approach to generate segmentation masks for hands and hand-held objects. Although the above methods form fine-grained localization of hands and objects, the lack of awareness of spatio-temporal information often results in temporal discontinuities when interactive objects are subject to deformation and occlusion.

\section{Method}
\begin{figure}
	\centering 
	\includegraphics[scale=0.99]{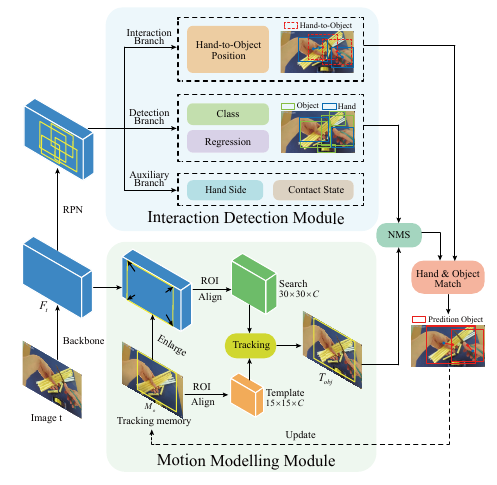}
	
	\caption{The architecture of our proposed TIO. It consists of two primary modules, namely the interaction detection module (indicated by blue block) and the motion modeling module (indicated by green block).     The former discovers the interacting objects from the scene by estimating the hand-to-object position and updates them to the memory. The latter is responsible for keeping track of the interacting objects in memory. 
    The two are combined to produce accurate and continuous interacting object trajectories in the sequence.
 }
 	\vspace{-15pt}
	\label{fig-framwork}
\end{figure}

Our proposed TIO aims to exploit spatio-temporal information to improve the localization of interactive objects in videos.
As shown in Figure~\ref{fig-framwork}, our TIO mainly consists of two modules, an interaction detection module that identifies interacting object by establishing the spatial relation with the hand and updates it to the tracking memory. 
A motion modeling module that keeps track of each interacting object in the memory by modeling the consistency and continuity of object appearance across frames.
These two modules use the same backbone to exact the feature from the image.
Given the image $I_t$ at frame $t$, we denote the image feature as $F_t$. The image feature and the detected interacting location of the previous frame which is stored in the tracking memory are denoted as $F_s$ and $M_o={\{m_i\}}_{i=1}^N$, $N$ denotes the number of objects.

\subsection{Discovering interacting objects via hand detection}
During hand-object interaction, distracting objects are frequently present in the scene along with the interacting objects and the interacting objects can be switched frequently in untrimmed videos. Thus, it is a challenge to identify the interactive objects and to update the tracking memory adaptively. To solve the problem, we propose to leverage hands information during the interaction to identify interacting objects as hands are relatively easy to detect and localize compared with unknown objects: AP of the hand detector reaches 90\% while that of objects lower than 50\%.  Specifically, to make use of the hand information, we design an interaction detection module (shown at the top of Fig. \ref{fig-framwork}) consisting of a detection branch for discovering hands and candidate objects in the scene, and an interaction branch for recognizing the interacting object from candidate objects based on the hand gesture.  

The detection branch is built upon Faster-RCNN \cite{fasterrcnn} object detector. 
For the $F_t$ extracted by the backbone, we use a region proposal network (RPN) to generate proposals. To obtain the locations of hands and potential interactive objects, we use a classification branch and a regression branch to predict the category and regression bounding box for each proposal. 
Different from the standard detectors, there are only three categories for classification prediction: hand, object, and background.

The interaction branch takes as input a proposal belonging to the hand and outputs an estimation of the location of the object with which it interacts. 
Let $B_h = (x_h, y_h, w_h, h_h)$ denote the bounding box of a hand $\mathcal{H}$, and let $B_o =(x_o, y_o, w_o, h_o)$ denote the ground truth bounding box of the object $\mathcal{O}$ with which it interacts.
The target corresponding to the interactive prediction $b_{o|h}$ is defined as:
\begin{equation}
\hat{b}_{o|h} = \{\frac{x_o - x_h}{w_h}, \frac{y_o - y_h}{h_h}, log \frac{w_o}{w_h}, log\frac{h_o}{h_h}\}.
\end{equation}
The training object of the interaction branch is to minimize the L1 loss between predictions and labels. Existed work \cite{100doh} predicts the center position of interacting objects to find the object being interacted with from the scene, which tends to be affected by heavy occlusion and distracting objects in adjacent centers. Our interaction branch infers the geometric information of the interacting object according to the gesture of the hand, which can be a more powerful cue to recognize interactive objects.

After obtaining the hand-to-object position estimation, we use it as an association cue to filter irrelevant objects.
For the hand $\mathcal{H}$, we calculate the compatibility scores between the estimated hand-to-object position $b_{o|h}$ and
the candidate objects $\mathcal{C}=\{c_j\}_{j=1}^{N_c}$ detected in the detection branch. $N_c$ is the number of candidates. The score is
\begin{equation}
score_{j} = exp(||b_{o|h} - b_{c_j|h}||),
\end{equation}
where $b_{c_j|h}$ represents the bounding box of $c_j$ with respect to $B_h$.
Intuitively, the $score_{j}$ indicates the probability of contact between the candidate object $c_j$ and the hand $\mathcal{H}$.
Based on the scores, we assign the best match to $\mathcal{H}$:
\begin{equation}
c_{j*} = arg max_{j=1}^{N_c} (score_{j}).
\end{equation}
We regard $c_{j*}$ as the object interacting with the $\mathcal{H}$.
With this design, we can discover interacting objects adaptively and then update them to the tracking memory for providing reliable template information.

To further understand the interaction activities, following \cite{100doh}, we add an extra branch to predict the hand side and contact state. 
This branch is trained by minimizing the cross entropy between the predictions and the ground truths.

\subsection{Continuous locating interactive objects by tracking} 
In this subsection, we will describe how to use the Siamese-based tracker (shown at the bottom of Fig. \ref{fig-framwork} to model the motion of an interacting object in successive frames.
We take the $k$-th interactive object in $M_o$, $m_k$,  as an example to illustrate how to create a motion association between adjacent frames.
Considering that the motion of the object is smooth, we determine the search region for $k$-th object on $I_t$ based on its historical location, rather than building a motion association over the entire image.
That is, we expand the width and height of $m_k$ by a factor of 2 to obtain $m_k^t$ as the search region.
Then we project $m_k$ and $m_k^t$ into $F_s$ and $F_t$ respectively to produce the template feature $T_k$ and the search region feature $S_k$. 
Since the shapes of $T_k$ and $S_k$ vary with the aspect ratio and size of the different interactive objects, they are unsuitable to be directly used as input for similarity matching. To solve this problem, we adopt an ROI align operation to obtain the template feature and search region feature, which are of fixed size. Formally,

\begin{equation}
\begin{aligned}
T_k=ROI Align(F_s, m_k),   \\
S_k=ROI Align(F_t, m_k^t),
\end{aligned}
\end{equation}
where $T_k \in \mathbb{R}^{15 \times 15 \times C}$ and $S_k \in \mathbb{R}^{30 \times 30 \times C}$.

\begin{figure}
\centering 
 	\includegraphics[scale=1.5]{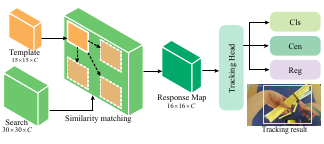}
	\caption{Illustration of the motion modeling module. It adopts a cross-correlation operation to perform similarity matching between the template and the search region. The generated response map is decoded to predict foreground probabilities "cls", the target center-ness score "cen" and the regression bounding box "reg".}
	\label{fig-track}
 	\vspace{-10pt}
\end{figure}

Effective propagation of the template to the search region is critical to localize the target. 
To balance tracking accuracy and speed, we adopt a cross-correlation operation to perform similarity matching 
between the $T_k$ and $S_k$ in Figure~\ref{fig-track}. Specifically, the whole $T_k$ is regarded as a convolutional kernel to perform a depth-wise correlation with $S_k$:
\begin{equation}
R_k=T_k * S_k,
\end{equation}
where $*$ represents the correlation operation. A response map $R_k \in \mathbb{R}^{16 \times 16 \times C}$ is generated that encodes the similarity relation of search reSiamese-based template.
Finally, we use the tracking head of SiamCAR \cite{SiamCAR} to decode the target state from the response map.
The outputs of the tracking are as follows:
\begin{equation}
\{A_{cls}, A_{cen}, A_{reg}\}=Dec(R_k),
\end{equation}
where the $A_{cls} \in \mathbb{R}^{16 \times 16 \times 2}$ represents the probabilities that pixels in $R_k$ belong to the foreground, $A_{cen} \in \mathbb{R}^{16 \times 16 \times 1}$ represents the probabilities that pixels in $R_k$ belong to the target center, and $A_{reg} \in \mathbb{R}^{16 \times 16 \times 4}$ represents the regressions for object bounding box estimation of all pixels.
We combine classification scores and center-ness scores to infer the object's central location and the corresponding regression prediction for that location as the bounding box of the target.

During the inference phase, when there are multiple interactive objects on the image, we can initialize multiple trackers for tracking in parallel. With the above design, we achieve continuous localization of the interactive object by modeling its motion across frames.

\subsection{Training and Inference}
Our proposed TIO consists of a tracking module and a detection module, both of which share a feature network. The full model is trained into two stages, in the first stage we train the backbone and interaction detection module with the following:
\begin{equation}
\mathcal{L}_{det}=\mathcal{L}_{rpn}+\mathcal{L}_{det\_cls}+\mathcal{L}_{det\_reg}+\mathcal{L}_{hand2obj}+L_{aux},
\end{equation}
where $\mathcal{L}_{rpn}$, $\mathcal{L}_{det\_cls}$ and $\mathcal{L}_{det\_reg}$ are the standard detection losses in Faster-RCNN, $\mathcal{L}_{hand2obj}$ is for training the interaction branch, and $\mathcal{L}_{aux}$ is for training the hand side and contact state predictions.
After finishing the first phase of training, we freeze the parameters of the backbone network and train the tracking module.
The training object of the motion modeling module can be formulated as:
\begin{equation} \label{track-loss}
\mathcal{L}_{motion}=\mathcal{L}_{cls} + \mathcal{L}_{cen}+ \mathcal{L}_{reg}. 
\end{equation}
$\mathcal{L}_{cls}$ is the cross-entropy loss used to train the classification branch, $L_{cen}$ is the BCE loss used to train the center-ness branch, and $\mathcal{L}_{reg}$ is the IOU loss used to train the regression branch.

During inference, when the interaction detection branch assigns a new interactive object to the hand, we add it to the tracking memory for subsequent tracking. And when the original tracking object is detected as no longer being contacted, we remove it from tracking memory to stop localization. Since the detection and tracking modules may produce duplicate bounding boxes for an object, we use a standard NMS operation on candidate objects by two modules before  assigning interactive objects.
It should be noted that we prioritize retaining the objects from the tracking module given that it is more continuous in time and space. 
With the above design, we can provide continuous localization results of the interactive objects, even if they are frequently switched.

\section{Experiments}

\subsection{Implementation Details}
We use a ResNet-100 \cite{resnet} as the backbone, which is pre-trained on ImageNet \cite{russakovsky2015imagenet}. We resize the input image to ensure that the longest side does not exceed 600 pixel, and take the output of ResNet-100 layer 3 as the image feature.

The interaction detection module is implemented using a standard Faster-RCNN. 
We use the training set of 100DOH \cite{100doh} to train the interaction detection module,
which contains approximately $100K$ images across $11$ daily interaction categories. The annotations in this dataset include hands' bounding boxes, hand sides, hand contact states, and objects' bounding boxes. Following the training strategy in 100DOH \cite{100doh}, we train our detection module with stochastic gradient descent (SGD) for $8$ epochs with batch size of $1$. The initial learning rate is $10^{-3}$ and decays by $0.1$ every $3$ epochs.

The motion modeling module is trained by pairs of images. However, it is no available video-level annotation in the current hand-object interaction benchmark. Therefore, we use GOT-10k \cite{got10k}, a large-scale, high-diversity benchmark for tracking objects in the wild, as the primary training set to train our tracking module. Interactive objects may suffer from heavy occlusion and deformation due to hand interaction. To enhance domain adaptation, we manually select about 9k image pairs  containing the same interactive object from 100DOH \cite{100doh} training set as supplementary training data.
We use SGD with momentum as the optimizer. We train our model for $25k$ iterations with batch size of 8. The initial learning rate is $0.02$ and decays by $0.1$ at $60\%$ and $80\%$ of iterations respectively.

\begin{table}[t]
    \setlength{\abovecaptionskip}{0.cm}
    \setlength{\belowcaptionskip}{2cm}	
    \begin{center}
    \caption{Evaluation of interactive object localization under different \textbf{scenes} on DOH-Trk. We compare our full model "$TIO$", a variant model "$TIO_{off}$" with the state-of-the-art method "$100DOH$" \cite{100doh}, using AP as an evaluation metric.
    }
    \vspace{10pt}
    \label{tab-scene}
    \setlength{\tabcolsep}{2.2mm}
    \begin{spacing}{1.25}
    \begin{tabular}{l|ccc}
        \toprule
        Scene (Num) &  $100DOH$ \cite{100doh} & $TIO_{off}$ & $TIO$  \\ \midrule
        Diy (249) & 45.89 & \textcolor{blue}{46.11} & \textcolor{red}{55.95}  \\
        Furniture (256) &  37.78 & \textcolor{blue}{38.19}  & \textcolor{red}{46.17}  \\
        Study (294) &  38.47 & \textcolor{blue}{45.39} & \textcolor{red}{47.13}  \\ 
        Repair (166) & 34.04 & \textcolor{blue}{36.73} & \textcolor{red}{40.03}  \\
        Packing (201) &  48.11  & \textcolor{blue}{52.61} & \textcolor{red}{53.27}  \\
        Puzzle (504) & 86.30 & \textcolor{blue}{87.33} & \textcolor{red}{87.44}  \\
        Gardening (156) & \textcolor{blue}{36.05} & 35.44 & \textcolor{red}{36.85}  \\ 
        Boardgame (149) & 40.27 & \textcolor{red}{45.98} & \textcolor{blue}{41.46}  \\ 
        Housework (181) & \textcolor{red}{40.78} & 39.76 & \textcolor{blue}{40.49}  \\ 
        Drink (166) & \textcolor{red}{43.24} & 40.52 & \textcolor{blue}{42.40}  \\ 
        Food (189) &  \textcolor{red}{36.14} & 35.33& \textcolor{blue}{35.79}  \\ 
        \hline
        Average (2511)  & 47.83 & \textcolor{blue}{49.75} & \textcolor{red}{50.27} \\
    \bottomrule
    \end{tabular}
    \end{spacing}
\end{center}
\vspace{-30pt}
\end{table}

\subsection{Test dataset}
To evaluate our proposed method, we curate a video-level dataset from 100DoH as there is no such annotated data available. 
We collect a video-level hand-object interaction dataset where part of frames are annotated by 100DOH \cite{100doh}.
Specifically, we randomly select annotated frames containing interactive actions from the 100DOH \cite{100doh} test subset, and download the corresponding videos from the Internet. For each video, we parse and extract the 60 images before the annotated frame. In this way, we collect 2511 test sequences with hand-object interactions, namely DOH-Trk.

\subsection{Evaluation protocol and metrics}
\begin{figure*}
\centering 
 	\includegraphics[scale=0.22]{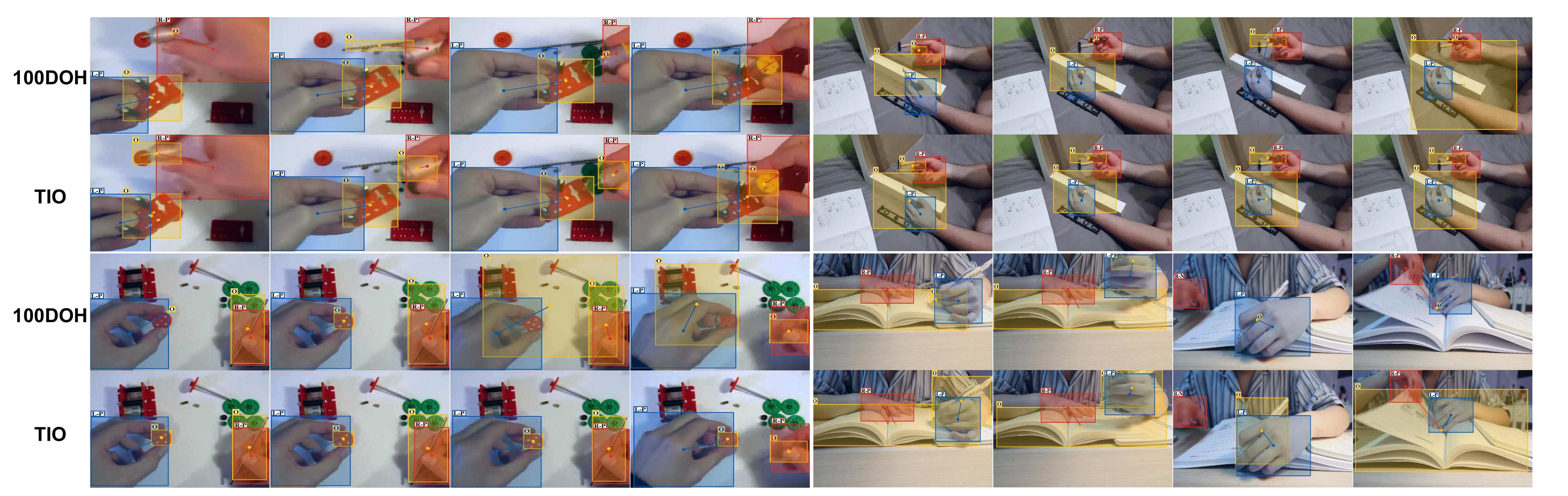}
	\caption{Comparisons of our TIO with the state-of-the-art method 100DOH \cite{100doh} on four sequences from DOH-Trk. Benefiting from the motion modeling over time, our TIO successfully handles the challenges such as heavy occlusion, background clutters, and distracting objects. }
	\label{fig-seqences}
 	\vspace{-15pt}
\end{figure*}

For the evaluation, the methods requiring temporal information run on the entire video sequence, but are evaluated only on frames with annotations; the frame-based methods are evaluated on the same annotated frames.
Following 100DOH \cite{100doh}, we adopt Average Precision (AP) as an evaluation indicator for quantitative comparisons, which is wildly used in object detection. 
For a detected hand, it is considered a true positive when the IOU with ground truth is greater than the thresh $\tau$ ($\tau$ = 0.5 in the experiment).  A detected object is a true positive only if it is successfully detected and matched with the correct hand. 
We provide the qualitative results of tracking results are shown in Figure~\ref{fig-seqences} and quantitative results in the following parts.

\subsection{Evaluation based on different scenes}
Depending on the interaction scene, the sequences in DOH-Trk can be divided into $11$ categories: diy, furniture, study, repair, packing, puzzle, gardening, boardgame, housework, drink, and food. We compare the performance of recognizing interactive objects with the state-of-the-art approach 100DOH \cite{100doh}. As shown in Table~\ref{tab-scene}, in the “diy” scenes with continuous interaction with different objects, we achieve an impressive improvement of $10.06\%$.  In the “furniture”, “study” and
“repair” scenes, our TIO also surpasses 100DOH \cite{100doh} by $8.38\%$, $8.66\%$, and $5.99\%$ respectively.
This proves that exploiting hand-object spatial relations and object appearance consistency can effectively improve the localization of interacting objects.
In addition, our variant model $"TIO_{off}"$ surpasses 100DOH \cite{100doh} by $6.92\%$, $3.5\%$ in the "study " and "packing" scenes respectively, which are only equipped with the motion modeling module. 
The results further demonstrate that modeling the motion of interacting objects is better able to cope with interference during the interaction than the frame-based detector.

\begin{table}
	\setlength{\abovecaptionskip}{0.cm}
	\setlength{\belowcaptionskip}{-0.4cm}	
	\begin{center}
         \caption{Evaluation of interactive object localization under different \textbf{contact states} on DOH-Trk. 
         We compare our "$TIO$" with the state-of-the-art method "$100DOH$" \cite{100doh}, using AP as an evaluation metric.
         }
 	\vspace{10pt}
	\label{tab-contact}
		\setlength{\tabcolsep}{4mm}
		\begin{spacing}{1.25}
        \begin{tabular}{l|cccc}
			\toprule
            Contact State & $100DOH$ \cite{100doh} &  $TIO$\\ 
            \midrule
            Self contact & 36.57  &  \textcolor{red}{37.22}  \\ 
            Other & 13.87 &  \textcolor{red}{18.4} \\ 
            Portable & 48.62 & \textcolor{red}{50.44} \\ 
            Non-portable & 26.32 & \textcolor{red}{34.05} \\ 
            \hline
            Average & 47.83 &  \textcolor{red}{50.27}\\ 
            \bottomrule
        \end{tabular}
		\end{spacing}
	\end{center}
	\vspace{-30pt}
\end{table}


\subsection{Evaluation based on different contact state}
The 100DOH dataset \cite{100doh} annotates four contact states of the hand, classifying the interaction objects as self, other person, portable object, and non-portable object. 
For the annotated frames in DOH-Trk, we divide them into four subsets according to the contact states, with subset sizes of 186, 14, 2295, and 229 respectively. 
As shown in Table~\ref{tab-contact}, we compare our proposed method with the state-of-the-art method 100DOH \cite{100doh}. 
Our proposed TIO improves the localization of interactive objects in four contact states. The AP score is improved from $47.83\%$ to $50.27\%$ in terms of overall contact states. It is noted that our TIO gains $7.73\%$ improvement from $26.32\%$ to $34.05\%$ on non-portable objects. The main reason is that non-portable objects in interactive scenes are often large and their perception is more easily affected by background clutter and distracting objects.
As a result, historical information more significantly enhances the state estimation of non-portable objects.


\begin{table}
	\setlength{\abovecaptionskip}{0.cm}
	\setlength{\belowcaptionskip}{-0.4cm}	
	\begin{center}

	\caption{Evaluation for full state prediction on DOH-Trk.}
         \vspace{10pt}
	\label{tab-full}
		\setlength{\tabcolsep}{1.5mm}
		\begin{spacing}{1.25}
        \begin{tabular}{l|ccccc}
        \toprule
            Method & Hand & H+State & H+Side & Object & All \\ 
            \midrule
            100DOH \cite{100doh} & \textcolor{red}{90.71} & \textcolor{red}{75.05} & \textcolor{red}{80.27} & 47.83 &  39.09 \\ 
            TIO & 90.69 & 74.11 & 80.15 & \textcolor{red}{50.27} & \textcolor{red}{45.4} \\  
            \bottomrule
        \end{tabular}
		\end{spacing}
	\end{center}
	\vspace{-30pt}
\end{table}

\begin{table}
	\setlength{\abovecaptionskip}{0.cm}
	\setlength{\belowcaptionskip}{-0.4cm}	
	\vspace{0pt}
	\begin{center}
	\caption{Comparisons of training dataset for motion modeling module.}
        \vspace{10pt}
	\label{tab-dataset}
		\setlength{\tabcolsep}{2mm}
		\begin{spacing}{1.25}
        \begin{tabular}{l|ccccc}
			\toprule
            Dataset & Object \\ 
            \midrule
            GOT-10k &  49.46 \\ 
            GOT-10k + 100DOH & 50.27  \\  
            \bottomrule
        \end{tabular}
		\end{spacing}
	\end{center}
	\vspace{-30pt}
 	\label{table-dataset}
\end{table}

\subsection{Evaluation of hand-object the full state perception}
We compare our TIO with the baseline for perceiving the full state of hand-object interaction, including hand location, hand side, hand contact state, and their corresponding interactive object.
The hand side and the contact state as auxiliary outputs of the detection module in our TIO.
Following 100DOH \cite{100doh}, we use AP as an evaluation metric to measure the performance of predicting hand state and hand side, and mAP as an evaluation metric to measure the performance of the full prediction “All”. A hand is considered true positive only if it has the correct side, contact state, and correct object associated with it. As shown in Table~\ref{tab-full}, compared with the 100DOH \cite{100doh}, our TIO surpasses it by $6.11\%$ for the "All".
The improvement of the perception of interacting objects facilitates the recognition of full-state interactions.
Since our detection module is trained on 100DOH dataset \cite{100doh}, the baseline adds egocentric images from \cite{Damen2018EPICKITCHENS, li2018eye, 2015-iccv-lending} as extra training dataset, the performance of hand side and hand contact state prediction is slightly lower than the baseline.

\subsection{Ablation Study}

\begin{figure}
\centering 
 	\includegraphics[scale=0.3]{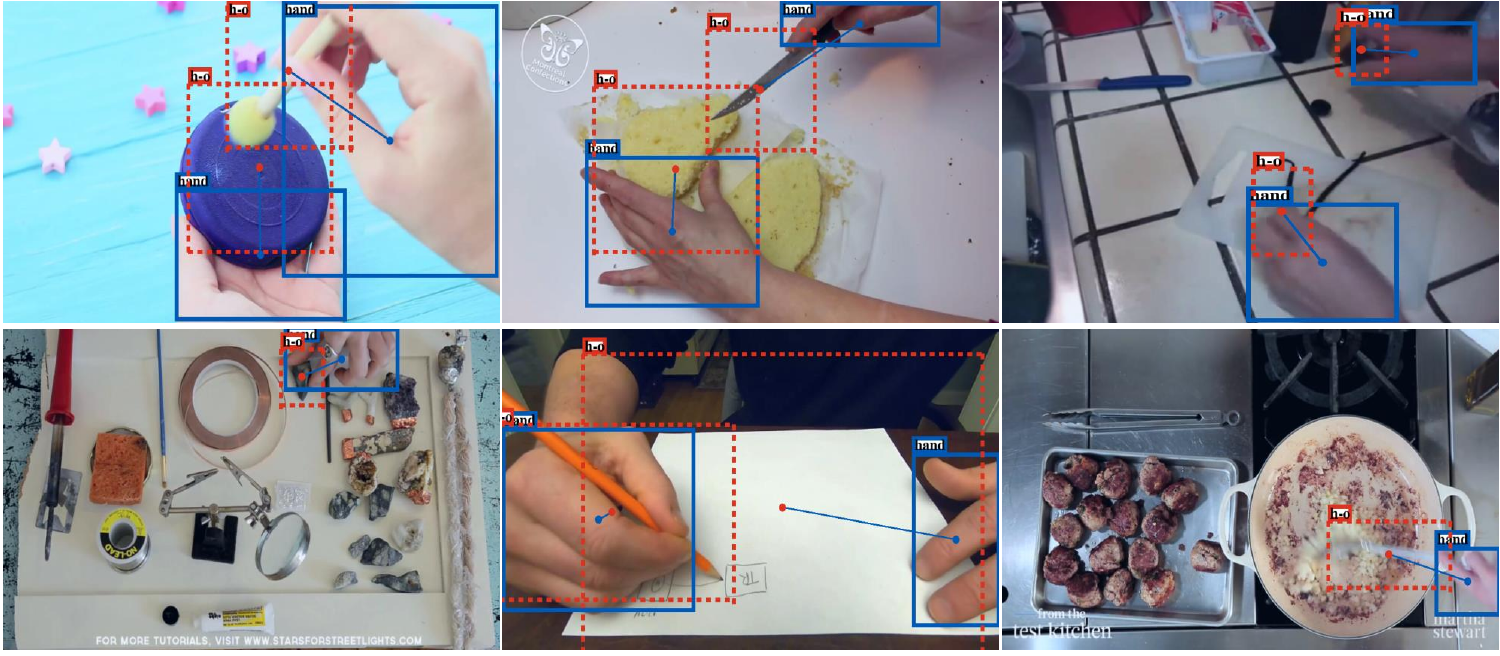}
	\caption{Estimating interacting object based on the appearance of the hand. The blue bounding boxes denote the detected hands, and the red dashed box denotes the predicted interacting objects. 
    }
    \vspace{-15pt}
	\label{fig-hand2object-vis}
\end{figure}

\textbf{Hand-to-object interaction branch.} 
To investigate the impact of the hand-to-object interaction branch, we equip our TIO with an offset branch proposed by 100DOH \cite{100doh}. This branch discovers the interacting object by predicting the offset between the hand and the object. 
As shown in Table~\ref{tab-scene}, when replaced with our proposed interaction branch, the AP score in "diy" is improved by $9.84\%$ from $46.11\%$ to $55.95\%$, the AP in "furniture" is improved by $7.98\%$ from $38.18\%$ to $46.16\%$.
This reason is that our proposed matching mechanism can better discover the interacting object by establishing hand-object spatial relations, thus providing more reliable template information for tracking.
Figure~\ref{fig-hand2object-vis} visualizes the bounding box of interactive objects predicted by the hand-to-object interaction branch. As we can see, the pose of the hand during interaction is closely related to the shape of the object.
Our method can predict the shape and position of the interaction object based on the appearance of the hand, thus providing a strong cue to match the correct objects.

\textbf{Training dataset for motion modeling.}
Despite the GOT-10k \cite{got10k} dataset containing various categories of objects. To further enhance domain adaptation, we manually collect 9k image pairs from 100DOH \cite{100doh} as a supplement to training our motion modeling module. 
To investigate the impact of training data, we train our TIO only with GOT-10k \cite{got10k} dataset. As shown in Table~\ref{table-dataset}, by adding image pairs from the hand-object dataset, the AP score is improved by $0.81\%$.

\section{CONCLUSIONS}
We present a novel object tracking framework to track interacting objects during hand-object interaction in the wild. 
To deal with the absence of interacting object priors, we design the interaction detection module that explores the hand-object spatial relation to discover the interacting object from the scenes. 
Subsequently, the motion modeling module is proposed to track interacting objects between successive frames, which can better cope with heavy occlusion, background clutters and distracting objects.
A video-level hand-object interaction dataset (DOH-Trk) is curated for testing and evaluation. 
The qualitative and quantitative results on DOH-Trk show that our method significantly improves the localization of interacting objects in the sequence.




{\small
\normalem
\bibliographystyle{IEEEtran}
\bibliography{IEEEfull}

\begin{thebibliography}{10}
\providecommand{\url}[1]{#1}
\csname url@samestyle\endcsname
\providecommand{\newblock}{\relax}
\providecommand{\bibinfo}[2]{#2}
\providecommand{\BIBentrySTDinterwordspacing}{\spaceskip=0pt\relax}
\providecommand{\BIBentryALTinterwordstretchfactor}{4}
\providecommand{\BIBentryALTinterwordspacing}{\spaceskip=\fontdimen2\font plus
\BIBentryALTinterwordstretchfactor\fontdimen3\font minus
  \fontdimen4\font\relax}
\providecommand{\BIBforeignlanguage}[2]{{%
\expandafter\ifx\csname l@#1\endcsname\relax
\typeout{** WARNING: IEEEtran.bst: No hyphenation pattern has been}%
\typeout{** loaded for the language `#1'. Using the pattern for}%
\typeout{** the default language instead.}%
\else
\language=\csname l@#1\endcsname
\fi
#2}}
\providecommand{\BIBdecl}{\relax}
\BIBdecl

\bibitem{karunratanakul2020grasping}
K.~Karunratanakul, J.~Yang, Y.~Zhang, M.~J. Black, K.~Muandet, and S.~Tang,
  ``Grasping field: Learning implicit representations for human grasps,'' in
  \emph{International Conference on 3D Vision}.\hskip 1em plus 0.5em minus
  0.4em\relax IEEE, 2020, pp. 333--344.

\bibitem{jiang2021hand}
H.~Jiang, S.~Liu, J.~Wang, and X.~Wang, ``Hand-object contact consistency
  reasoning for human grasps generation,'' in \emph{IEEE/CVF International
  Conference on Computer Vision}, 2021, pp. 11\,107--11\,116.

\bibitem{2022-eccv-dexmv}
Y.~Qin, Y.-H. Wu, S.~Liu, H.~Jiang, R.~Yang, Y.~Fu, and X.~Wang, ``Dexmv:
  Imitation learning for dexterous manipulation from human videos,'' in
  \emph{European Conference on Computer Vision}, 2022, pp. 570--587.

\bibitem{2021-icra-learning}
P.~Mandikal and K.~Grauman, ``Learning dexterous grasping with object-centric
  visual affordances,'' in \emph{{IEEE} Journal of Robotics and Automation},
  2021, pp. 6169--6176.

\bibitem{2019-cvpr-contextual}
S.~Narasimhaswamy, Z.~Wei, Y.~Wang, J.~Zhang, and M.~Hoai, ``Contextual
  attention for hand detection in the wild,'' in \emph{IEEE/CVF Conference on
  Computer Vision and Pattern Recognition}, 2019, pp. 9567--9576.

\bibitem{2015-iccv-lending}
S.~Bambach, S.~Lee, D.~J. Crandall, and C.~Yu, ``Lending a hand: Detecting
  hands and recognizing activities in complex egocentric interactions,'' in
  \emph{IEEE/CVF International Conference on Computer Vision}, 2015, pp.
  1949--1957.

\bibitem{2018-tip-Joint}
X.~Deng, Y.~Zhang, S.~Yang, P.~Tan, L.~Chang, Y.~Yuan, and H.~Wang, ``Joint
  hand detection and rotation estimation using cnn,'' \emph{{IEEE} Transactions
  on Image Processing}, vol.~27, no.~4, pp. 1888--1900, 2018.

\bibitem{2018-cvpr-lifestyle}
D.~F. Fouhey, W.-c. Kuo, A.~A. Efros, and J.~Malik, ``From lifestyle vlogs to
  everyday interactions,'' in \emph{IEEE/CVF Conference on Computer Vision and
  Pattern Recognition}, 2018, pp. 4991--5000.

\bibitem{100doh}
D.~Shan, J.~Geng, M.~Shu, and D.~F. Fouhey, ``Understanding human hands in
  contact at internet scale,'' in \emph{IEEE/CVF Conference on Computer Vision
  and Pattern Recognition}, 2020, pp. 9866--9875.

\bibitem{zhou2022apptracker}
T.~Zhou, W.~Luo, Z.~Shi, J.~Chen, and Q.~Ye, ``Apptracker: Improving tracking
  multiple objects in low-frame-rate videos,'' in \emph{Proceedings of the 30th
  ACM International Conference on Multimedia}, 2022, pp. 6664--6674.

\bibitem{luo2019end}
W.~Luo, P.~Sun, F.~Zhong, W.~Liu, T.~Zhang, and Y.~Wang, ``End-to-end active
  object tracking and its real-world deployment via reinforcement learning,''
  \emph{{IEEE} Transactions on Pattern Analysis and Machine Intelligence},
  vol.~42, no.~6, pp. 1317--1332, 2019.

\bibitem{luo2021multiple}
W.~Luo, J.~Xing, A.~Milan, X.~Zhang, W.~Liu, and T.-K. Kim, ``Multiple object
  tracking: A literature review,'' \emph{Artificial intelligence}, vol. 293, p.
  103448, 2021.

\bibitem{siamrpn++}
B.~Li, W.~Wu, Q.~Wang, F.~Zhang, J.~Xing, and J.~Yan, ``Siamrpn++: Evolution of
  siamese visual tracking with very deep networks,'' in \emph{IEEE/CVF
  Conference on Computer Vision and Pattern Recognition}, 2019.

\bibitem{SiamCAR}
D.~Guo, J.~Wang, Y.~Cui, Z.~H. Wang, and S.~Chen, ``Siamcar: Siamese fully
  convolutional classification and regression for visual tracking,'' in
  \emph{IEEE/CVF Conference on Computer Vision and Pattern Recognition}, 2020.

\bibitem{9839629}
H.~Pu, L.~He, P.~Cheng, M.~Sun, and J.~Chen, ``Security of industrial robots:
  Vulnerabilities, attacks, and mitigations,'' \emph{{IEEE} Network}, vol.~37,
  no.~1, pp. 111--117, 2023.

\bibitem{9810792}
S.~He, K.~Shi, C.~Liu, B.~Guo, J.~Chen, and Z.~Shi, ``Collaborative sensing in
  internet of things: A comprehensive survey,'' \emph{{IEEE} Communications
  Surveys and Tutorials}, vol.~24, no.~3, pp. 1435--1474, 2022.

\bibitem{liu2021semi}
S.~Liu, H.~Jiang, J.~Xu, S.~Liu, and X.~Wang, ``Semi-supervised 3d hand-object
  poses estimation with interactions in time,'' in \emph{IEEE/CVF Conference on
  Computer Vision and Pattern Recognition}, 2021, pp. 14\,687--14\,697.

\bibitem{cai2019exploiting}
Y.~Cai, L.~Ge, J.~Liu, J.~Cai, T.-J. Cham, J.~Yuan, and N.~M. Thalmann,
  ``Exploiting spatial-temporal relationships for 3d pose estimation via graph
  convolutional networks,'' in \emph{IEEE/CVF International Conference on
  Computer Vision}, 2019, pp. 2272--2281.

\bibitem{Chen2020JointH3}
Y.~Chen, Z.~Tu, D.~Kang, R.~Chen, L.~Bao, Z.~Zhang, and J.~Yuan, ``Joint
  hand-object 3d reconstruction from a single image with cross-branch feature
  fusion,'' \emph{{IEEE} Transactions on Image Processing}, vol.~30, pp.
  4008--4021, 2020.

\bibitem{2021-cvpr-reconstructing}
Z.~Cao, I.~Radosavovic, A.~Kanazawa, and J.~Malik, ``Reconstructing hand-object
  interactions in the wild,'' in \emph{IEEE/CVF International Conference on
  Computer Vision}, 2021, pp. 12\,417--12\,426.

\bibitem{2022-cvpr-human}
M.~Goyal, S.~Modi, R.~Goyal, and S.~Gupta, ``Human hands as probes for
  interactive object understanding,'' in \emph{IEEE/CVF Conference on Computer
  Vision and Pattern Recognition}, 2022, pp. 3293--3303.

\bibitem{li2013pixel}
C.~Li and K.~M. Kitani, ``Pixel-level hand detection in ego-centric videos,''
  in \emph{IEEE/CVF Conference on Computer Vision and Pattern Recognition},
  2013, pp. 3570--3577.

\bibitem{2022-cvpr-forward}
M.~Huang, S.~Narasimhaswamy, S.~Vazir, H.~Ling, and M.~Hoai, ``Forward
  propagation, backward regression, and pose association for hand tracking in
  the wild,'' in \emph{IEEE/CVF Conference on Computer Vision and Pattern
  Recognition}, 2022, pp. 6406--6416.

\bibitem{2020-nips-detecting}
S.~Narasimhaswamy, T.~Nguyen, and M.~H. Nguyen, ``Detecting hands and
  recognizing physical contact in the wild,'' \emph{Neural Information
  Processing Systems}, vol.~33, pp. 7841--7851, 2020.

\bibitem{misra2017red}
I.~Misra, A.~Gupta, and M.~Hebert, ``From red wine to red tomato: Composition
  with context,'' in \emph{IEEE/CVF Conference on Computer Vision and Pattern
  Recognition}, 2017, pp. 1792--1801.

\bibitem{naeem2021learning}
M.~F. Naeem, Y.~Xian, F.~Tombari, and Z.~Akata, ``Learning graph embeddings for
  compositional zero-shot learning,'' in \emph{IEEE/CVF Conference on Computer
  Vision and Pattern Recognition}, 2021, pp. 953--962.

\bibitem{nagarajan2018attributes}
T.~Nagarajan and K.~Grauman, ``Attributes as operators: factorizing unseen
  attribute-object compositions,'' in \emph{European Conference on Computer
  Vision}, 2018, pp. 169--185.

\bibitem{fathi2013modeling}
A.~Fathi and J.~M. Rehg, ``Modeling actions through state changes,'' in
  \emph{IEEE/CVF Conference on Computer Vision and Pattern Recognition}, 2013,
  pp. 2579--2586.

\bibitem{liu2017jointly}
Y.~Liu, P.~Wei, and S.-C. Zhu, ``Jointly recognizing object fluents and tasks
  in egocentric videos,'' in \emph{IEEE/CVF International Conference on
  Computer Vision}, 2017, pp. 2924--2932.

\bibitem{2020-eccv-forecasting}
M.~Liu, S.~Tang, Y.~Li, and J.~M. Rehg, ``Forecasting human-object interaction:
  joint prediction of motor attention and actions in first person video,'' in
  \emph{European Conference on Computer Vision}, 2020, pp. 704--721.

\bibitem{2021-cvpr-anticipative}
R.~Girdhar and K.~Grauman, ``Anticipative video transformer,'' in
  \emph{IEEE/CVF Conference on Computer Vision and Pattern Recognition}, 2021,
  pp. 13\,505--13\,515.

\bibitem{2022-eccv-fine}
L.~Zhang, S.~Zhou, S.~Stent, and J.~Shi, ``Fine-grained egocentric hand-object
  segmentation: Dataset, model, and applications,'' in \emph{European
  Conference on Computer Vision}, 2022, pp. 127--145.

\bibitem{2021-nips-cohesiv}
D.~Shan, R.~Higgins, and D.~Fouhey, ``Cohesiv: Contrastive object and hand
  embedding segmentation in video,'' \emph{Neural Information Processing
  Systems}, vol.~34, pp. 5898--5909, 2021.

\bibitem{fasterrcnn}
S.~Ren, K.~He, R.~Girshick, and J.~Sun, ``Faster r-cnn: Towards real-time
  object detection with region proposal networks,'' in \emph{Neural Information
  Processing Systems}, C.~Cortes, N.~Lawrence, D.~Lee, M.~Sugiyama, and
  R.~Garnett, Eds., vol.~28.\hskip 1em plus 0.5em minus 0.4em\relax Curran
  Associates, Inc., 2015.

\bibitem{resnet}
K.~He, X.~Zhang, S.~Ren, and J.~Sun, ``Deep residual learning for image
  recognition,'' in \emph{IEEE/CVF Conference on Computer Vision and Pattern
  Recognition}, 2016.

\bibitem{russakovsky2015imagenet}
O.~Russakovsky, J.~Deng, H.~Su, J.~Krause, S.~Satheesh, S.~Ma, Z.~Huang,
  A.~Karpathy, A.~Khosla, and M.~Bernstein, ``Imagenet large scale visual
  recognition challenge,'' \emph{International Journal of Computer Vision},
  2015.

\bibitem{got10k}
L.~Huang, X.~Zhao, and K.~Huang, ``Got-10k: A large high-diversity benchmark
  for generic object tracking in the wild,'' \emph{{IEEE} Transactions on
  Pattern Analysis and Machine Intelligence}, 2018.

\bibitem{Damen2018EPICKITCHENS}
D.~Damen, H.~Doughty, G.~M. Farinella, S.~Fidler, A.~Furnari, E.~Kazakos,
  D.~Moltisanti, J.~Munro, T.~Perrett, W.~Price, and M.~Wray, ``Scaling
  egocentric vision: The epic-kitchens dataset,'' in \emph{European Conference
  on Computer Vision}, 2018.

\bibitem{li2018eye}
Y.~Li, M.~Liu, and J.~M. Rehg, ``In the eye of beholder: Joint learning of gaze
  and actions in first person video,'' in \emph{European Conference on Computer
  Vision}, 2018, pp. 619--635.

\end{thebibliography}
}

\end{document}